\documentclass[11pt]{article}

\usepackage[letterpaper,margin=1in]{geometry}
\usepackage{amsmath,amssymb,amsfonts}
\usepackage{newtxtext,newtxmath}
\usepackage{microtype}
\usepackage{graphicx}
\usepackage{xcolor}
\usepackage{booktabs}
\usepackage{array}
\usepackage{multirow}
\usepackage{enumitem}
\usepackage{caption}
\usepackage{listings}
\usepackage{fancyvrb}
\usepackage{float}
\usepackage{xeCJK}
\setCJKsansfont{FandolHei-Regular.otf}
\setCJKmonofont{FandolFang-Regular.otf}
\floatstyle{plain}
\newfloat{listing}{t}{lop}
\floatname{listing}{Listing}
\usepackage{tikz}
\usetikzlibrary{positioning,arrows.meta,fit,backgrounds,calc,shapes.geometric}
\usepackage[colorlinks=true,linkcolor=blue!60!black,citecolor=blue!60!black,urlcolor=blue!60!black]{hyperref}
\hypersetup{
  pdftitle={Voyag: Joint Multi-Intent Detection and Execution-Relation Classification for In-Vehicle Voice Assistants},
  pdfauthor={SEN LUO},
  pdfsubject={Multi-intent detection, execution relation, calibration, in-vehicle dialogue},
  pdfkeywords={multi-intent detection; execution relation; resource conflict; joint training; calibration; in-vehicle dialogue}
}

\definecolor{voyblue}{HTML}{1F4E79}
\definecolor{voygreen}{HTML}{2E7D32}
\definecolor{voypurple}{HTML}{6A1B9A}
\definecolor{voygrey}{HTML}{F2F2F2}

\lstdefinestyle{jsonsample}{
  basicstyle=\ttfamily\scriptsize,
  breaklines=true,
  columns=fullflexible,
  keepspaces=true,
  frame=single,
  rulecolor=\color{black!30},
  backgroundcolor=\color{voygrey},
  framexleftmargin=4pt,
  xleftmargin=4pt,
  aboveskip=8pt,
  belowskip=8pt,
}

\newcommand{\voyag}{\textsc{Voyag}}
\newcommand{\mbert}{mmBERT-base}
\DeclareMathOperator*{\argmax}{arg\,max}
\DeclareMathOperator{\softmax}{softmax}
\newcommand{\R}{\mathbb{R}}

\title{\vspace{-1.2em}\textbf{\voyag{}: Joint Multi-Intent Detection and Execution-Relation\\ Classification for In-Vehicle Voice Assistants}\vspace{0.4em}}
\author{SEN LUO\\ Launch Tech Co., Ltd.\\ \texttt{597582250@qq.com}}
\date{}

\begin{document}
\maketitle
\thispagestyle{empty}

% ---------------------------------------------------------------------
\begin{abstract}
\noindent
An in-vehicle voice assistant must decide not only \emph{which} intents an utterance contains but also \emph{how} co-active intents may be executed: concurrently, in a fixed order, or not at all without arbitration. We present \voyag{}, a single-forward decision model that jointly predicts ten controlled vehicle-intent slots and a three-way execution relation (parallel, sequential, mutually exclusive) over each co-active intent pair. \voyag{} combines a multilingual ModernBERT-style encoder (\mbert{}, 306.9\,M parameters), a typed per-slot choice head, and a lightweight GroupCalibration relation head (1.61\,M parameters) that consumes marker hidden states, pooled context, explicit eight-dimensional resource-requirement profiles, and symmetrized outer-product conflict features. All objectives---soft cross-entropy over slot choices, focal cross-entropy over relations, and a mutex-consistency penalty---are optimized in one joint loss. On a synthetic Chinese vehicle-life benchmark with semantic scene-level splits and a leakage audit (30{,}017 / 3{,}473 / 5{,}000 groups), validation-only model selection yields 0.9893 slot accuracy and 0.0526 expected calibration error over 50{,}000 slot decisions; the intent head is left unscaled and is mildly under-confident. The relation head temperature is fitted on validation only ($T=4.5517$, 2{,}773 pairs), yielding test ECE 0.0550. Conditioned on gold intents, the relation head reaches 0.8548 accuracy, 0.8473 macro-F1 and 0.6062 mutually-exclusive recall over 4{,}000 co-active pairs; end to end, driven only by its own intent predictions in one encoder forward, it reaches 0.8530 accuracy and 0.8464 macro-F1 (95\% bootstrap CI 0.842--0.864). Against a genuine two-stage rule pipeline consuming the same predicted intents, that end-to-end gain is small and only borderline ($+1.3$ points, 95\% CI $-0.3$ to $+2.9$, bootstrap $p=0.05$), so we claim no aggregate accuracy advantage; the learned head's advantage is concentrated where surface rules are structurally blind, in sequential recall (0.961 vs.\ 0.769) and parallel recall (0.993 vs.\ 0.753). We report deterministic rule baselines with an explicit tautology disclosure, a cross-iteration case study in which relation-only fine-tuning destroyed the intent head, and remaining failures, most notably a held-out thermal conflict pair on which end-to-end accuracy is 0.106.
\end{abstract}

\vspace{0.4em}

\vspace{1.2em}
% ---------------------------------------------------------------------
\section{Introduction}
\label{sec:intro}

Consider two Chinese in-vehicle commands: \emph{``open the window and set the air conditioning to 24 degrees''} and \emph{``play music and answer the incoming call''}. Both contain exactly two active intents, yet their execution policies differ fundamentally. The first pair touches disjoint resources and can be dispatched concurrently. The second pair contends for the same audio-output channel, and a safe assistant must clarify before acting. Conventional pipelines detect the intent set first and encode execution policy afterwards, in external rule tables that grow with the ontology and remain disconnected from the decision model.

We argue that execution relation is a \emph{decision-layer} quantity and should be predicted by the same forward pass that produces the intents. \voyag{} turns this claim into a concrete model. An utterance and a serialized vehicle state enter one shared \mbert{} encoder together with ten controlled slot questions; a typed choice head decodes every slot, and a GroupCalibration relation head scores each co-active pair into \{parallel, sequential, mutually exclusive\}. The relation head is deliberately small (1.61\,M parameters, 0.50\% of the checkpoint) and is attached with a zero-initialized residual so that it cannot perturb the base decision model at initialization.

The main contributions are as follows.
\begin{enumerate}[leftmargin=1.6em,itemsep=2pt,topsep=4pt]
  \item \textbf{Joint supervision.} Intent, relation, and mutex-consistency losses are optimized simultaneously. We document a comparison of two training iterations (Section~\ref{sec:ablation}): the relation-only objective used in the earlier iteration, with an unfrozen encoder tail, collapsed slot accuracy from 0.88 to 0.51 while leaving relation accuracy superficially acceptable. The two iterations differ in data lineage as well as in objective, so we present this as a case study rather than as a controlled ablation.
  \item \textbf{Explicit resource semantics.} Rather than hoping a hidden state encodes resource contention, the relation head receives an eight-dimensional resource-requirement profile per action and symmetrized outer-product conflict features (72 dimensions), making the mutex decision readable and auditable.
  \item \textbf{Audited synthetic evaluation.} Labels come from a closed-form resource ontology; splits are disjoint at the semantic-scene level; a near-duplicate leakage audit passes; model selection uses validation only and the test split is evaluated exactly once. Because labels are ontology-derived, our deterministic rule baselines are reported as \emph{reference upper bounds} with an explicit tautology disclosure, not as independent competitors.
\end{enumerate}

On the V8 benchmark, \voyag{} reaches 0.9893 slot accuracy over 50{,}000 slot decisions and 0.0526 expected calibration error (ECE). Conditioned on gold intents it scores 0.8548 relation accuracy, 0.8473 macro-F1 and 0.6062 recall on the mutually-exclusive class over 4{,}000 pairs; end to end, on its own intent predictions and still one encoder forward per utterance, it scores 0.8530 relation accuracy and 0.8464 macro-F1 (95\% bootstrap CI 0.842--0.864). The dominant remaining failure is generalization to a held-out thermal conflict pair (\texttt{window.open} $\times$ \texttt{hvac.defrost}), on which end-to-end accuracy is 0.106; we analyze it in Section~\ref{sec:errors}.

% ---------------------------------------------------------------------
\section{Related Work}
\label{sec:related}

\paragraph{Multi-intent detection.} Prior work detects multiple intents per utterance via slot--intent graphs, semantic embedding matching, or large-language-model prompting over synthetic data~\cite{glgin,agif}. Closer to our setting, Gangadharaiah and Narayanaswamy cast multiple-intent detection and slot labeling as one joint prediction problem for goal-oriented dialogue and report that modelling the two together beats a sequential pipeline, because slot evidence disambiguates which of several candidate intents are actually live~\cite{gangadharaiah2019}. Deployed assistants converged on the same joint, on-device formulation for latency and privacy reasons rather than for accuracy alone~\cite{snips}. All of these methods stop at the intent set: none predicts whether two detected intents may execute together, which is the quantity an in-vehicle runtime actually has to commit to.

\paragraph{Intent dependency and ordering.} Relation annotation has a long history in temporal reasoning. MATRES separates intent, aspect and temporal relation onto distinct axes and shows that a multi-axis scheme sharply improves inter-annotator agreement over a single conflated relation label~\cite{matres}; our execution relation is a deliberately narrow analogue of that idea, restricted to pairs that are co-active within one utterance and grounded in a resource ontology rather than in time. At the other end of the design space, LLM agent frameworks recover order and preconditions at inference time---by interleaving reasoning traces with tool actions~\cite{react}, by translating a request into a task plan dispatched to specialist models~\cite{hugginggpt}, or by prompting for an explicit plan before solving~\cite{plansolve}. These are expressive but pay for it in latency, cost and auditability, and none emits a calibrated confidence per relation. \voyag{} instead fixes a three-class pairwise relation vocabulary and predicts it discriminatively inside a low-latency decision layer, trading open-ended planning for auditable, low-latency arbitration.

\paragraph{Narrow calibrated decision models.} Compact encoder-plus-head models remain attractive for on-device assistants because they offer bounded latency and calibrated confidences over controlled vocabularies~\cite{guo2017calibration}. \voyag{} extends this paradigm with a pairwise relation head and a joint objective, and contributes a documented case study of when multi-head fine-tuning preserves or destroys a previously learned head.

\paragraph{Resource-conflict modeling.} Mutex detection in our benchmark is generated from an explicit resource ontology (audio output, microphone, display, thermal management, drivetrain, cabin envelope). Exposing this ontology as input features to the relation head follows the principle of hybrid neuro-symbolic design: learnable weights operate on human-auditable predicates~\cite{neurosymbolic}.

% ---------------------------------------------------------------------
\section{Task Definition}
\label{sec:task}

\subsection{Slots, options, and groups}
Let $\mathcal{S}=\{s_1,\dots,s_{10}\}$ be the controlled slots: window, hvac, media, navigation, seat, light, phone\_call, charging, message, diagnostic. Each slot $s$ has a closed option set $\mathcal{O}_s$ that always contains a \texttt{none} option (e.g.\ $\mathcal{O}_{\text{hvac}}=\{\texttt{none},\texttt{set\_temperature},\texttt{increase\_fan},\texttt{decrease\_fan},\texttt{defrost}\}$). An example is a tuple $(u, \sigma, Q)$ of utterance $u$, vehicle state $\sigma$, and the ten slot questions $Q=\{q_s\}_{s\in\mathcal{S}}$. The intent target is $y_s\in\mathcal{O}_s$ for every slot; the set of active slots is $\mathcal{A}=\{s: y_s\neq\texttt{none}\}$.

\subsection{Execution relations}
For each unordered active pair $(i,j)\in\mathcal{A}\times\mathcal{A},\, i<j$, the target relation is $r_{ij}\in\{0,1,2\}$:
\begin{itemize}[leftmargin=1.6em,itemsep=1pt,topsep=3pt]
  \item $0$ \textbf{parallel}: both actions may execute concurrently;
  \item $1$ \textbf{sequential}: execution requires a fixed order (dependency);
  \item $2$ \textbf{mutually exclusive}: the actions contend for an irreconcilable resource and require arbitration or user clarification.
\end{itemize}
Every relation-supervised group in V8 has exactly $|\mathcal{A}|=2$ active slots and therefore contributes exactly one labelled pair; groups with $|\mathcal{A}|\le 1$ carry intent supervision only. This is why intent and relation item counts differ (50{,}000 vs.\ 4{,}000 at test), and it bounds the scope of the relation head: it is never supervised on groups of three or more co-active intents, so its behaviour on larger groups is undefined rather than merely untested (Section~\ref{sec:limitations}).

\subsection{Resource ontology and conflict predicate}
Every controlled action $a$ carries a binary resource-requirement vector $p(a)\in\{0,1\}^{8}$ over the axes
\begin{equation}
\label{eq:axes}
[\,\underbrace{p_0}_{\text{audio out}},\ \underbrace{p_1}_{\text{mic}},\ \underbrace{p_2}_{\text{display}},\ \underbrace{p_3}_{\text{climate active}},\ \underbrace{p_4}_{\text{drivetrain}},\ \underbrace{p_5}_{\text{envelope open}},\ p_6,\ p_7\,].
\end{equation}
The last two axes, $p_6$ and $p_7$, are currently unassigned and reserved for future resource types. The six active axes correspond to audio output, microphone, display, climate-active, drivetrain, and envelope-open.
Two actions $a,b$ conflict iff at least one resource predicate fires:
\begin{equation}
\label{eq:conflict}
\begin{aligned}
\mathrm{audio}(a,b)   &= p_0(a)\land p_0(b),\\
\mathrm{display}(a,b) &= p_2(a)\land p_2(b),\\
\mathrm{driving}(a,b) &= p_4(a)\land p_4(b),\\
\mathrm{thermal}(a,b) &= \big(p_3(a)\land p_5(b)\big)\lor\big(p_5(a)\land p_3(b)\big),
\end{aligned}
\end{equation}
and $r_{ij}=2$ iff any predicate in \eqref{eq:conflict} holds. The thermal predicate is cross-diagonal: an active thermal management action conflicts with opening the cabin envelope. In V8, \texttt{hvac.set\_temperature} is defined as climate-\emph{inactive} ($p_3=0$), so \texttt{window.open} $\times$ \texttt{hvac.set\_temperature} is parallel, whereas \texttt{window.open} $\times$ \texttt{hvac.defrost} and \texttt{window.open} $\times$ \texttt{seat.heat\_on} remain mutually exclusive. Sequential labels come from a curated dependency list (e.g.\ \texttt{diagnostic.read\_dtc} before \texttt{seat.heat\_on}); all remaining active pairs are parallel.

% ---------------------------------------------------------------------
\section{Model}
\label{sec:model}

\subsection{Grouped single-forward serialization}
\label{sec:serialization}
All ten slot questions of one example are concatenated into a single sequence so that one encoder forward serves every slot and the relation head:
\begin{equation}
\label{eq:serial}
[\texttt{CLS}]\ \bigoplus_{s\in\mathcal{S}}\Big([\texttt{TYPE}]\, q_s^{\text{instr}}\, [\texttt{SEP}]\, \bigoplus_{o\in\mathcal{O}_s} m_{s,o}\ o\ [\texttt{SEP}]\Big)\ \oplus\ \tau(\sigma)\ [\texttt{SEP}],
\end{equation}
where $m_{s,o}$ is a dedicated marker token in front of every option description and $\tau(\sigma)$ is the vehicle state, tokenized \emph{once} and shared by all questions. Marker positions and a marker mask are carried alongside the sequence so that option-level hidden states can be gathered after the encoder. Two sequence budgets appear in the training and inference configurations and should not be conflated: training uses $\texttt{max\_len}=1024$, as recorded in \texttt{weights/rl\_agent\_config.json}, whereas the exported ONNX graph and the published schema fix $\texttt{seq\_len}=512$ for on-device inference. With the per-question head limit $\texttt{head\_max\_len}=256$, the longest serialized V8 row encodes to 209 tokens, so neither budget truncates any example and the two configurations are numerically equivalent on this data.

\subsection{Shared encoder and typed choice head}
The encoder is \mbert{} (\texttt{jhu-clsp/mmBERT-base}; \cite{modernbert,mmBERT}), a ModernBERT-style multilingual encoder: 22 transformer layers, hidden size 768, 12 attention heads, vocabulary 256{,}000, RoPE with $\theta=160{,}000$, and alternating local (window 128) / global attention every third layer; it holds 306{,}939{,}648 parameters. For question $q_s$ and option $o$, the marker hidden state $h_{s,o}$ is pooled into a choice logit $z_{s,o}$ by the existing typed decision head (type embedding, transformer head, LayerNorm, linear scorer). The per-slot prediction is $\hat y_s=\argmax_{o\in\mathcal{O}_s} z_{s,o}$, and the winning marker state $h_{s,\hat y_s}$ is the intent state passed downstream.

\subsection{GroupCalibration relation head}
\label{sec:groupcalib}
Let $x_i = h_{i,\hat y_i} + e_{g(i)} \in \R^{768}$ be the intent state of active slot $i$, augmented by a learned group embedding $e_g$, and let $\tilde x_i = f_{\text{proj}}(x_i)$ be its LayerNorm--linear--GELU--dropout projection. For an active pair $(i,j)$ the relation head consumes
\begin{equation}
\label{eq:phi}
\phi(i,j)=\big[\ \tilde x_i,\ \tilde x_j,\ |\tilde x_i-\tilde x_j|,\ \tilde x_i\odot\tilde x_j,\ \bar c,\ \pi_i,\ \pi_j,\ \operatorname{sym}(p_i p_j^{\top}),\ \min(p_i,p_j)\ \big],
\end{equation}
where $\bar c\in\R^{768}$ is the mean pooled context of the group, $\pi_i\in\R^{8}$ is a projected resource profile, $\operatorname{sym}(p_i p_j^{\top}) = p_i p_j^{\top} + p_j p_i^{\top}\in\R^{64}$ is the symmetrized outer product over the \emph{raw} requirement vectors, and $\min(p_i,p_j)$ is their elementwise minimum. Hence $\phi(i,j)\in\R^{5\cdot 768 + 2\cdot 8 + 72}=\R^{3928}$. The symmetrized outer product is order-invariant and encodes both same-resource contention (diagonal terms, e.g.\ two audio streams) and cross-resource opposition (off-diagonal terms, e.g.\ climate-active vs.\ envelope-open), i.e.\ a differentiable relaxation of the predicates in \eqref{eq:conflict}. A two-layer MLP maps $\phi(i,j)$ to three relation logits:
\begin{equation}
\label{eq:relmlp}
r(i,j) = W_2\,\mathrm{Dropout}\big(\mathrm{GELU}(W_1 \phi(i,j) + b_1)\big) + b_2,\qquad W_1\in\R^{256\times 3928},\ W_2\in\R^{3\times 256}.
\end{equation}
During training, whole intent-state vectors of a pair are dropped with a tuned probability (\emph{state dropout}), forcing the head to rely on the auditable resource/conflict features and pooled context instead of memorizing pair-specific hidden patterns.

The same module also performs residual candidate calibration,
\begin{equation}
\label{eq:rescal}
\hat z_{s,o} = z_{s,o} + \big(w^{\top} f_{\text{proj}}(x_{s,o}) + b\big)\cdot \mathbf{1}[o \text{ valid}],\qquad w=\mathbf{0},\ b=0 \text{ at init},
\end{equation}
so attaching the module is exactly identity at initialization, and $W_2,b_2$ in \eqref{eq:relmlp} are likewise zero-initialized. GroupCalibration adds 1{,}611{,}868 parameters (0.50\% of the 323{,}520{,}866-parameter checkpoint).

\subsection{Joint objective}
\label{sec:objective}
\paragraph{Intent term.} With gold option distribution $y_{s}\in\Delta^{|\mathcal{O}_s|}$ (a smoothed one-hot, e.g.\ 0.96 on the target option) and option mask $M_s$,
\begin{equation}
\label{eq:intent}
\mathcal{L}_{\text{intent}} = -\frac{1}{|\mathcal{S}|}\sum_{s\in\mathcal{S}} \sum_{o\in\mathcal{O}_s} y_{s,o}\, M_{s,o}\, \log \softmax_o\!\big(z_{s}\odot M_s\big).
\end{equation}
\paragraph{Relation term.} Over valid pairs $\mathcal{P}$ with focal modulation~\cite{focal} and class weights that up-weight the mutex class,
\begin{equation}
\label{eq:focal}
\mathcal{L}_{\text{rel}} = \frac{\sum_{(i,j)\in\mathcal{P}} w_{r_{ij}}\,(1-p_{ij,r_{ij}})^{\gamma}\,(-\log p_{ij,r_{ij}})}{\sum_{(i,j)\in\mathcal{P}} w_{r_{ij}}},\qquad p_{ij}=\softmax(r(i,j)).
\end{equation}
\paragraph{Mutex-consistency term.} For a pair labelled mutually exclusive, both members should not look simultaneously executable. With per-candidate confidence $\hat p_{s}=\max_o \softmax(\hat z_s)_o$,
\begin{equation}
\label{eq:mutex}
\mathcal{L}_{\text{mutex}} = \mathbb{E}_{(i,j):\, r_{ij}=2}\big[\min(\hat p_i, \hat p_j)\big],
\end{equation}
a smooth penalty that leaves a strongly preferred side intact while suppressing two nearly equal high confidences.
\paragraph{Total.}
\begin{equation}
\label{eq:total}
\mathcal{L} = \lambda_i\, \mathcal{L}_{\text{intent}} + \lambda_r\, \mathcal{L}_{\text{rel}} + \lambda_m\, \mathcal{L}_{\text{mutex}},
\end{equation}
Every batch contributes \eqref{eq:intent} even when it contains no active pair, so the trainable part of the encoder continues to receive intent supervision instead of drifting under relation-only pressure.

\begin{figure*}[t]
\centering
\begin{tikzpicture}[
  font=\small, >=Stealth, node distance=6mm,
  block/.style={draw=black!60, fill=white, rounded corners=2pt, minimum height=8mm, align=center, inner sep=5pt, text width=11cm},
  io/.style={draw=voyblue, fill=voyblue!8, rounded corners=2pt, minimum height=8mm, align=center, inner sep=5pt, text width=8cm},
  enc/.style={draw=voygreen!70!black, fill=voygreen!8, rounded corners=2pt, minimum height=9mm, align=center, inner sep=5pt, text width=11cm},
  head/.style={draw=voypurple!70!black, fill=voypurple!8, rounded corners=2pt, minimum height=9mm, align=center, inner sep=5pt, text width=5.9cm},
  loss/.style={draw=black!50, fill=black!5, rounded corners=2pt, minimum height=8mm, align=center, inner sep=5pt, text width=9cm},
  wide/.style={draw=black!60, fill=white, rounded corners=2pt, minimum height=8mm, align=center, inner sep=5pt, text width=12.8cm},
  lbl/.style={font=\scriptsize, text=black!60},
  every node/.append style={execute at begin node={\hyphenpenalty=10000 \exhyphenpenalty=10000}},
]
\node[io] (in) {utterance $u$ + vehicle state $\sigma$ + 10 controlled slot questions};
\node[block, below=of in] (ser) {grouped serialization, Eq.~\eqref{eq:serial}: one \texttt{input\_ids} / \texttt{attention\_mask} / \texttt{marker\_mask} batch};
\node[enc, below=of ser] (enc) {shared \mbert{} encoder: 22 ModernBERT layers, 306.9\,M parameters\\ \scriptsize one forward per utterance ($\texttt{encoder\_calls}=1$)};
\node[block, below=of enc] (gather) {marker-state gather: option states $h_{s,o}$; intent states $x_i$ for active slots; pooled context $\bar c$};
\node[head, below=10mm of gather, xshift=-42mm] (choice) {typed choice head\\ $\hat y_s=\argmax_o z_{s,o}$ over 10 slots};
\node[head, below=10mm of gather, xshift=42mm] (gc) {GroupCalibration, 1.61\,M params\\ $\phi(i,j)$ Eq.~\eqref{eq:phi} $\to$ MLP Eq.~\eqref{eq:relmlp}};
\node[wide, below=26mm of gather] (out) {decision assembly: intents, pairwise relations, mutex violations, execution hint $\in$ \{single, parallel, sequential, clarify\}};
\node[loss, below=8mm of out] (loss) {joint objective, Eq.~\eqref{eq:total}: $\mathcal{L}_{\text{intent}} + \mathcal{L}_{\text{rel}} + \lambda_m\,\mathcal{L}_{\text{mutex}}$};
\draw[->] (in) -- (ser);
\draw[->] (ser) -- (enc);
\draw[->] (enc) -- (gather);
\draw[->] (gather.south -| choice.north) -- (choice.north);
\draw[->] (gather.south -| gc.north) -- (gc.north);
\draw[->] (choice.south) -- (choice.south |- out.north);
\draw[->] (gc.south) -- (gc.south |- out.north);
\draw[->] (out) -- (loss);
\draw[->,dashed] (loss.west) -- ++(-3.0,0) node[lbl, above, pos=0.5] {one backward pass} |- (ser.west);
\end{tikzpicture}
\caption{Overview of \voyag{}. A single grouped forward through one shared encoder produces all ten slot decisions and every co-active pair relation; the joint objective back-propagates through both heads simultaneously.}
\label{fig:overview}
\end{figure*}
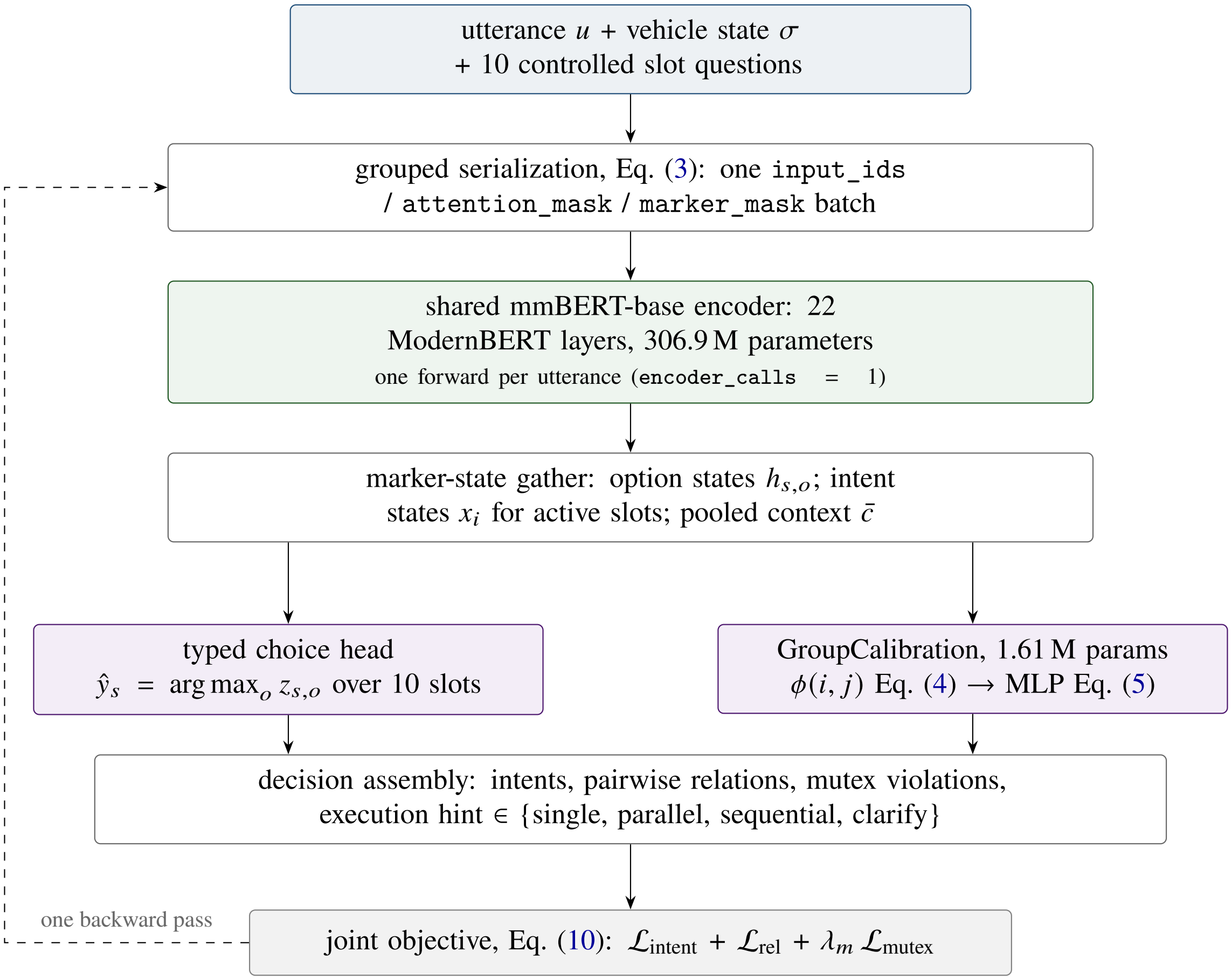

% ---------------------------------------------------------------------
\section{The V8 Benchmark}
\label{sec:data}

\subsection{Generation and validation}
V8 is a synthetic Chinese vehicle-life benchmark. Scenes are produced by a deterministic rule generator grounded in the resource ontology of Section~\ref{sec:task}, complemented by LLM-assisted scene generation; every generated row passes strict deterministic validation (valid schema, two active slots, unique non-empty utterances, both intents covered, no contradictory relation assignments). Surface realizations are templated with lexical, syntactic, and ordering variation, including negation, paraphrases, implicit requests, out-of-domain rejection, and hard negatives. Every row is validated against the schema before entering a split.

\subsection{Splits and leakage audit}
Splits are disjoint at the \emph{semantic-scene} level: a complete intent tuple (e.g.\ \texttt{sequential:\allowbreak diagnostic.\allowbreak read\_dtc\allowbreak|seat.\allowbreak heat\_on}) belongs to exactly one split, so no scene appears under different surface forms across splits. Table~\ref{tab:splits} summarizes the splits; the audit reports zero scene overlap, zero semantic-scene overlap, zero duplicate ids and utterances, and zero near-duplicate violations at char-3-gram cosine threshold 0.82 (maximum observed cross-split similarity 0.8182). Relation supervision exists only for active--active pairs; single-intent and zero-intent groups balance the intent task. The thermal mutex family is split so that \texttt{window.open} $\times$ \texttt{hvac.defrost} appears \emph{only} in test, providing a held-out conflict-generalization probe.

\begin{table}[t]
\centering
\caption{V8 split composition (groups; relation rows in parentheses). Train relation rows are balanced 1:1:1.}
\label{tab:splits}
\footnotesize
\begin{tabular}{lrrrr}
\toprule
 & train & val & test & total\\
\midrule
groups & 30{,}017 & 3{,}473 & 5{,}000 & 38{,}490\\
active--active pairs & 24{,}007 & 2{,}773 & 4{,}000 & 30{,}780\\
single-intent groups & 3{,}006 & 350 & 500 & 3{,}856\\
zero-intent groups & 3{,}004 & 350 & 500 & 3{,}856\\
semantic scenes & 193 & 21 & 30 & 244\\
\midrule
parallel / sequential / mutex & 8{,}007/8{,}000/8{,}000 & 933/933/907 & 1{,}334/1{,}333/1{,}333 & --\\
mutex mix (aud/disp/driv/therm) & 1600/5600/534/266 & 467/440/0/0 & 445/444/0/444 & --\\
\bottomrule
\end{tabular}
\end{table}

% ---------------------------------------------------------------------
\section{Experimental Setup}
\label{sec:setup}

\paragraph{Initialization.} Training starts from an intent-healthy base checkpoint (slot accuracy 0.8801 on the V7 test split), whose choice head was trained with intent supervision only.

\paragraph{Optimization.} We use AdamW with differential learning rates for the relation head and the unfrozen encoder tail, gradient clipping, and mixed precision. A small number of encoder tail layers plus the final norm are opened together with GroupCalibration: 21{,}672{,}796 trainable parameters (6.7\% of the checkpoint). Class weighting and focal modulation encourage the harder mutex class, and state dropout regularizes the pair representation.

\paragraph{Selection protocol.} Model selection uses \emph{validation} macro-F1 with a mutex-recall floor of 0.40: an epoch whose validation mutex recall falls below the floor can never win; among epochs above the floor, the highest macro-F1 wins. The test split is evaluated exactly once, after selection. Relation confidences are temperature-scaled for reporting~\cite{guo2017calibration}; the scalar $T=4.5517$ is fitted on the 2{,}773 validation relation pairs only and then applied to the test split, yielding mean confidence 0.8623 and ECE 0.0550. At $T=1$ those test diagnostics are mean confidence 0.9863 and ECE 0.1321. Because temperature scaling is monotone per row, all argmax-based metrics (accuracy, precision, recall, F1) are invariant to it. No temperature is fitted for the intent head.

\paragraph{Evaluation backends.} Intent metrics and the gold-intent-conditioned relation metrics are produced by the PyTorch checkpoint. The end-to-end numbers in Section~\ref{sec:rel-results} are produced by the exported ONNX graph on CUDA, which is the artefact used for on-device inference. The two backends agree: over the same 50{,}000 test slot decisions the ONNX graph scores 0.98932 against 0.9893 for the PyTorch evaluation, i.e.\ the same value at the precision we report, and a per-output comparison on fixed inputs matches to $7\times10^{-6}$ on all four graph outputs. We therefore treat the two as interchangeable, quote the PyTorch figure as the headline slot accuracy, and state the backend wherever an end-to-end number appears.

% ---------------------------------------------------------------------
\section{Results}
\label{sec:results}

\subsection{Intent detection and calibration}
\label{sec:intent-results}
Table~\ref{tab:slots} reports per-slot accuracy over 50{,}000 test slot decisions (5{,}000 per slot). Overall slot accuracy is 0.9893 with ECE 0.0526~\cite{guo2017calibration}, mean confidence 0.9455, and Brier score 0.0165. No slot collapses; the weakest slot, media (0.8998), is analyzed in Section~\ref{sec:errors}.

The sign of the residual miscalibration matters for deployment. Mean confidence (0.9455) sits \emph{below} accuracy (0.9893), so the choice head is mildly under-confident rather than over-confident; a runtime that thresholds on confidence will over-clarify slightly rather than commit to wrong actions. We fit no post-hoc temperature on the intent head, in contrast to the relation head, whose reporting temperature is fitted on validation only (Section~\ref{sec:setup}).

\begin{table}[t]
\centering
\caption{Per-slot intent accuracy on the V8 test split (5{,}000 decisions per slot).}
\label{tab:slots}
\small
\begin{tabular}{lrlrlr}
\toprule
slot & acc & slot & acc & slot & acc\\
\midrule
window & 0.9998 & light & 1.0000 & message & 1.0000\\
hvac & 0.9986 & phone\_call & 0.9994 & charging & 1.0000\\
media & 0.8998 & seat & 0.9956 & diagnostic & 1.0000\\
navigation & 1.0000 & & & \textbf{overall} & \textbf{0.9893}\\
\bottomrule
\end{tabular}
\end{table}

\subsection{Execution-relation classification}
\label{sec:rel-results}
We report the relation head under two conditioning regimes. They answer different questions, and conflating them is a common source of inflated numbers in this area, so we keep them separate throughout.

\paragraph{Gold-intent conditioned.} Table~\ref{tab:rel} scores the relation head on the 4{,}000 test pairs whose two intents are taken from the labels. This isolates relation-head quality from intent error and is the regime in which the rule baselines of Section~\ref{sec:baselines} are also scored. \voyag{} reaches 0.8548 accuracy and 0.8473 macro-F1. The mutually-exclusive class shows the expected precision--recall trade-off induced by the focal weight and the mutex-consistency penalty: precision 0.9417 at recall 0.6062. Sequential is recall-dominated (0.9655) at precision 0.7461, i.e.\ the head prefers sequential over parallel when order cues are ambiguous, and prefers parallel or sequential over mutex when conflict evidence is implicit.

\paragraph{End to end.} Table~\ref{tab:e2e} reports the deployed setting: one forward pass, no gold information, the runtime scoring only those pairs it believes are active-active. \voyag{} reaches 0.8530 accuracy and 0.8464 macro-F1, a 0.18-point drop from the gold-conditioned figure. Of the 4{,}000 supervised pairs, 3{,}992 receive an emitted relation and 8 do not, because the runtime did not regard both intents as active; those 8 are counted as errors and are deliberately \emph{not} credited to any relation class. Restricting to the 3{,}989 pairs whose two intents were both predicted correctly recovers 0.8551 accuracy, which localizes most of the end-to-end gap on the intent side rather than the relation side.

A 2{,}000-draw bootstrap over test rows gives a 95\% interval of $[0.842, 0.864]$ for end-to-end accuracy and $[0.835, 0.857]$ for macro-F1. Paired against the two-stage R2 pipeline consuming the same predicted intents, the accuracy difference is $+0.0130$ with 95\% CI $[-0.0032, +0.0285]$ and a one-sided bootstrap $p$ of 0.05. We therefore do not claim a clear aggregate end-to-end accuracy advantage over the rule pipeline; the effect is borderline under this resampling. This is worth stating plainly: because the mutex labels are produced by a closed-form predicate that R2 also evaluates, the rule enjoys an accuracy floor on this benchmark that a learned head is unlikely to beat in aggregate.

The learned head's advantage is qualitative rather than aggregate, and it lies where surface rules are structurally blind. R2 can call a pair sequential only when the utterance carries an explicit order cue, and otherwise defaults to parallel; \voyag{} reaches 0.9610 sequential recall against R2's 0.7689, and 0.9925 parallel recall against 0.7534. In the other direction R2 attains 0.9977 mutex recall against \voyag{}'s 0.6054, but only constructively. The practical reading is that the two are complementary rather than competitive: a deployable system should take mutex from the auditable resource predicate and take order and concurrency from the relation head, which additionally emits a temperature-scaled per-pair confidence that a rule cannot.

\begin{table}[t]
\centering
\caption{End-to-end relation classification on the V8 test split (4{,}000 supervised pairs; ONNX graph, CUDA). The model uses only its own intent predictions. A supervised pair the runtime declines to score is counted as wrong and is not credited to any relation class. The R2-pipeline block is the same surface heuristic driven by the same predicted intents.}
\label{tab:e2e}
\small
\begin{tabular}{lrrrr}
\toprule
class & support & precision & recall & F1\\
\midrule
\multicolumn{5}{l}{\emph{\voyag{} V8 end to end: accuracy 0.8530, macro-F1 0.8464}}\\
parallel & 1{,}334 & 0.9344 & 0.9925 & 0.9626\\
sequential & 1{,}333 & 0.7456 & 0.9610 & 0.8397\\
mutually exclusive & 1{,}333 & 0.9417 & 0.6054 & 0.7370\\
macro & 4{,}000 & 0.8739 & 0.8530 & 0.8464\\
\midrule
\multicolumn{5}{l}{\emph{R2 pipeline on predicted intents: accuracy 0.8400, macro-F1 0.8401}}\\
parallel & 1{,}334 & 0.7637 & 0.7534 & 0.7585\\
sequential & 1{,}333 & 0.7570 & 0.7689 & 0.7629\\
mutually exclusive & 1{,}333 & 1.0000$^\dagger$ & 0.9977 & 0.9989\\
\bottomrule
\multicolumn{5}{l}{}\\
\multicolumn{5}{l}{\scriptsize \voyag{} end-to-end confusion, rows target / columns prediction, order}\\
\multicolumn{5}{l}{\scriptsize parallel / sequential / mutex / not emitted:}\\
\multicolumn{5}{c}{\scriptsize $\begin{pmatrix} 1324 & 1 & 9 & 0\\ 5 & 1281 & 41 & 6\\ 88 & 436 & 807 & 2 \end{pmatrix}$}\\
\multicolumn{5}{l}{\scriptsize $^\dagger$ constructive upper bound (labels derive from the same ontology).}\\
\end{tabular}
\end{table}

\begin{table}[t]
\centering
\caption{Gold-intent-conditioned relation classification on the V8 test split (4{,}000 pairs; PyTorch checkpoint). The two intents of every pair are taken from the labels, so this table isolates relation-head quality from intent error. Confusion matrix rows: target, columns: prediction, order parallel / sequential / mutex.}
\label{tab:rel}
\small
\begin{tabular}{lrrrr}
\toprule
class & support & precision & recall & F1\\
\midrule
parallel & 1{,}334 & 0.9344 & 0.9925 & 0.9626\\
sequential & 1{,}333 & 0.7461 & 0.9655 & 0.8417\\
mutually exclusive & 1{,}333 & 0.9417 & 0.6062 & 0.7376\\
\midrule
macro & 4{,}000 & 0.8741 & 0.8547 & 0.8473\\
\bottomrule
\multicolumn{5}{c}{}\\
\multicolumn{5}{c}{confusion $=\begin{pmatrix} 1324 & 1 & 9\\ 5 & 1287 & 41\\ 88 & 437 & 808 \end{pmatrix}$}\\
\end{tabular}
\end{table}

\subsection{Epoch selection trace}
Table~\ref{tab:epochs} shows the validation trace used for selection. Epoch 5 wins (macro-F1 0.9266, mutex recall 0.9118). Validation loss is not monotone because the focal relation term dominates late-epoch gradients while intent loss stays flat near 0.273; the floor-guarded macro-F1 criterion is therefore the operative selector.

\begin{table}[t]
\centering
\caption{Validation trace (2{,}773 pairs). Selected epoch in bold.}
\label{tab:epochs}
\small
\begin{tabular}{rrrrr}
\toprule
epoch & train $\mathcal{L}$ & val acc & val macro-F1 & val mutex recall\\
\midrule
1 & 0.5120 & 0.8777 & 0.8765 & 0.7354\\
2 & 0.3097 & 0.8489 & 0.8448 & 0.6362\\
3 & 0.2882 & 0.9066 & 0.9060 & 0.8401\\
4 & 0.2803 & 0.8821 & 0.8813 & 0.7894\\
\textbf{5} & \textbf{0.2776} & \textbf{0.9268} & \textbf{0.9266} & \textbf{0.9118}\\
6 & 0.2766 & 0.8990 & 0.8979 & 0.7773\\
7 & 0.2765 & 0.9131 & 0.9124 & 0.8247\\
8 & 0.2709 & 0.9131 & 0.9125 & 0.9283\\
\bottomrule
\end{tabular}
\end{table}

\subsection{Baselines with tautology disclosure}
\label{sec:baselines}
Because V8 relation labels are generated from the closed-form ontology of Section~\ref{sec:task}, any rule that evaluates the same predicates reproduces the mutex class by construction. We therefore report two deterministic rules as \emph{reference upper bounds} and disclose the tautology:
\begin{itemize}[leftmargin=1.6em,itemsep=1pt,topsep=3pt]
  \item \textbf{R1 (closed-form reference)}: mutex iff \eqref{eq:conflict} fires; sequential iff the ordered pair is in the curated dependency list; else parallel.
  \item \textbf{R2 (surface heuristic)}: mutex iff \eqref{eq:conflict} fires; sequential iff the utterance contains an order cue (然后, 之后, 先, 再, 顺序, 随后, 接着, 接下来, 完成, 依次); else parallel. R2 does not consult the curated list and is the primary baseline.
\end{itemize}
Table~\ref{tab:baselines} compares the two rules with \voyag{} under gold-intent conditioning, and adds the un-fine-tuned base intent head as an intent-only reference. Under that conditioning \voyag{} exceeds R2 by 1.4 accuracy points (0.8548 vs.\ 0.8408) and 0.7 macro-F1 points, and exceeds the base intent head by 12.3 intent-accuracy points (0.9893 vs.\ 0.8662). Mutex recall stays well below the constructive rule bound, for the reason analyzed in Section~\ref{sec:errors}.

We deliberately omit the ``base intent head $+$ R2'' combination that an earlier draft of this work reported as a pipeline result. The R2 half of that combination was computed on gold intents, so the row paired a non-oracle intent accuracy with an oracle relation accuracy and overstated what a two-stage system would actually achieve. The genuine two-stage pipeline, in which R2 consumes the model's predicted intents, is reported in Table~\ref{tab:e2e} and discussed in Section~\ref{sec:rel-results}. Note that R2's parallel and sequential decisions depend only on surface order cues and not on which intents were resolved, so its parallel and sequential recalls are identical under both conditionings; only its mutex column changes.

\begin{table}[t]
\centering
\caption{Relation baselines on the V8 test split under gold-intent conditioning. R1/R2 mutex recall of 1.0 is constructive (same ontology) and is an upper bound, not an independent result. End-to-end rows, in which the relation stage consumes predicted intents, are given in Table~\ref{tab:e2e}.}
\label{tab:baselines}
\small
\begin{tabular}{lrrrrr}
\toprule
method & intent acc & rel acc & par.\ recall & seq.\ recall & mutex recall\\
\midrule
R1 closed-form reference & -- & 0.8333 & 1.0000 & 0.4996 & 1.0000$^\dagger$\\
R2 surface heuristic & -- & 0.8408 & 0.7534 & 0.7689 & 1.0000$^\dagger$\\
base intent head, intent only & 0.8662 & -- & -- & -- & --\\
\voyag{} V8 & \textbf{0.9893} & \textbf{0.8548} & \textbf{0.9925} & \textbf{0.9655} & 0.6062\\
\bottomrule
\multicolumn{6}{l}{\scriptsize $^\dagger$ constructive upper bound (labels derive from the same ontology).}\\
\end{tabular}
\end{table}

\subsection{Objective comparison across two training iterations}
\label{sec:ablation}
Table~\ref{tab:ablation} compares two training iterations of the same architecture and the same GroupCalibration head. In the V7 iteration the head was trained with a \emph{relation-only} objective ($\lambda_i=0$) while the last four encoder layers were unfrozen: relation accuracy stayed acceptable (0.848) but slot accuracy collapsed to 0.5143, because the shared encoder tail drifted without intent supervision and destroyed the typed choice head. Restoring \eqref{eq:intent} in every batch (V8) recovers and surpasses the base checkpoint's intent accuracy (0.9893 vs.\ 0.8801 base, 0.8662 base-on-V8-test) while improving relation accuracy over both rules.

We state the strength of this evidence explicitly. The V7 row is scored on the V7 test split and the V8 rows on the V8 test split, and the two iterations also differ in scene coverage and surface realization, so the objective is not the only variable that changed. The comparison is therefore a documented case study of what happened when we changed the objective, not a controlled ablation at fixed data; a matched-data ablation would require retraining the relation-only variant on the V8 split, which we did not do. The direction and the magnitude are nevertheless large (a 37-point loss in slot accuracy with relation accuracy essentially unchanged), so we regard the qualitative conclusion as well supported: multi-head fine-tuning of a shared encoder is unsafe unless intent supervision is present in every batch.

\begin{table}[t]
\centering
\caption{Objective comparison across two training iterations (same architecture, same head). The V7 row is scored on the V7 iteration test split; the V8 rows use the V8 test split. Data lineage is not held constant, so this is a case study rather than a controlled ablation.}
\label{tab:ablation}
\small
\begin{tabular}{llrr}
\toprule
objective & encoder tail & intent acc & rel acc\\
\midrule
relation only ($\lambda_i=0$) & unfrozen 4 layers & 0.5143 & 0.8480\\
intent only (base ckpt) & frozen & 0.8662 & --\\
joint, Eq.~\eqref{eq:total} & unfrozen 4 layers & \textbf{0.9893} & \textbf{0.8548}\\
\bottomrule
\end{tabular}
\end{table}

\subsection{Single-forward verification}
An instrumented runtime counts encoder invocations during grouped inference. Across four packaged verification cases (parallel thermal-neutral pair, mutex audio pair, parallel media/navigation pair, and the held-out thermal pair), the counter reports $\texttt{encoder\_calls}=1$ per utterance while producing all ten slot argmax decisions and every pair relation, confirming that the architecture is genuinely single-forward rather than a per-slot ensemble.

% ---------------------------------------------------------------------
\section{Error Analysis}
\label{sec:errors}

\paragraph{Held-out thermal conflict.} Of 525 mutex test errors under gold-intent conditioning, 437 are sequential predictions and 88 parallel. The held-out pair \texttt{window.open} $\times$ \texttt{hvac.defrost} (444 test pairs, absent from train and validation by design) accounts for the majority: without training exposure to the thermal predicate on this pair, the head defaults to the more frequent parallel/sequential geometry. This is a deliberate generalization probe, and it is the main reason mutex recall (0.6062) trails the constructive rule bound.

Decomposing the same failure by conflict family under end-to-end scoring makes its size explicit. Accuracy is 0.977 on display mutex (444 pairs), 0.733 on audio mutex (445 pairs) and 0.106 on the held-out thermal family (444 pairs). Of the 526 mutex-target pairs missed end to end, 397 are thermal, 119 audio and 10 display. The thermal family is therefore close to a total failure rather than a partial one, and removing it would raise end-to-end mutex recall from 0.605 to roughly 0.86. We report the unadjusted figure because the held-out family is part of the test split by construction, and because a model that cannot transfer a cross-diagonal resource predicate to an unseen action pair has not demonstrated conflict generalization in any useful sense.

A related caveat concerns model selection. The validation split contains only audio (467) and display (440) mutex pairs: thermal mutex is absent from validation entirely, and drivetrain mutex appears only in training (534 pairs) and in neither validation nor test. The epoch selector therefore never observed the conflict family that dominates the test failures, and no early-stopping signal could have flagged it.

\paragraph{Media commentary coverage.} Test zero-intent rows include music commentary (e.g.\ ``this song sounds nice'') labelled \texttt{media=none}. V8 train under-covers such commentary, so the media slot absorbs 501 of the 534 total slot errors (accuracy 0.8998). The labels are correct; the gap is coverage, not calibration.

\paragraph{Sequential over-prediction.} The confusion matrix is asymmetric in an interpretable way: of 1{,}334 parallel-target pairs only 10 are misread (1 sequential, 9 mutex); of 1{,}333 sequential-target pairs 46 are misread (5 parallel, 41 mutex); but of 1{,}333 mutex-target pairs 525 are misread (88 parallel, 437 sequential). Ambiguous dependency phrasing without explicit order cues, and implicit conflict evidence, remain the hardest regions, consistent with the precision/recall asymmetry in Table~\ref{tab:rel}.

% ---------------------------------------------------------------------
\section{Limitations}
\label{sec:limitations}
\begin{enumerate}[leftmargin=1.6em,itemsep=2pt,topsep=4pt]
  \item \textbf{Synthetic benchmark.} All numbers are from synthetic Chinese data; they are not real-user, real-vehicle, or on-device production metrics.
  \item \textbf{Homogeneous surface forms.} 4{,}982 of the 5{,}000 test rows carry a rule-templated surface style and only 18 are LLM-surfaced. Paraphrase and disfluency robustness is therefore barely probed, and the intent accuracy in particular should be read as an upper bound for real transcribed speech.
  \item \textbf{Pairs only.} Every relation-supervised group has exactly two active intents, so the relation head has never seen a group of three or more co-active intents and its behaviour on larger groups is undefined rather than merely untested.
  \item \textbf{Constructive rule bound.} Rule mutex recall of 1.0 derives from the same ontology that generated the labels; it is an upper bound for reference, not evidence that rules suffice in the wild.
  \item \textbf{No clear aggregate end-to-end gain.} Against a two-stage rule pipeline on the same predicted intents, the end-to-end accuracy difference is $+0.0130$ with 95\% CI $[-0.0032, +0.0285]$ and bootstrap $p=0.05$. Our claim is limited to the class-level asymmetry of Section~\ref{sec:rel-results}, not to aggregate accuracy.
  \item \textbf{Held-out conflict generalization.} The thermal pair probe fails; conflict generalization to unseen action combinations is unresolved.
  \item \textbf{No feature ablation.} The relation head consumes five heterogeneous feature groups (Eq.~\eqref{eq:phi}). We do not report a leave-one-group-out study, so the relative contribution of the projected intent states, the pooled context, the resource profiles and the symmetrized outer product is not quantified; the state-dropout rate was fixed during earlier development and not re-tuned on V8.
  \item \textbf{No matched-data objective ablation.} The V7/V8 comparison of Section~\ref{sec:ablation} changes objective and data lineage together. Retraining the relation-only variant on the V8 split, via \texttt{--intent-weight 0}, is the obvious next experiment and we regard it as the most important missing control in this paper.
  \item \textbf{State-conditioned inference.} Inputs include a serialized vehicle state; integrations must supply real state in the canonical field order.
  \item \textbf{No on-device latency study.} Single-forward is verified by instrumentation in Python; Android/edge latency is future work.
\end{enumerate}

% ---------------------------------------------------------------------
\section{Conclusion}
\label{sec:conclusion}
\voyag{} places execution-relation arbitration inside a narrow decision layer: one grouped encoder forward yields ten controlled intents and every co-active pair relation, trained end-to-end under a joint intent--relation--mutex objective. The cross-iteration comparison indicates that joint supervision is necessary rather than optional: without it, relation fine-tuning silently destroyed the intent head. We are equally explicit about what the results do not show. End to end, the model does not demonstrate a clear aggregate accuracy advantage over a two-stage rule pipeline, and it fails outright on a held-out thermal conflict family; its observed advantage is confined to order and concurrency decisions that surface rules cannot reach, and to the temperature-scaled per-pair confidence it emits alongside them. With audited synthetic data, explicit resource predicates, validation-only selection, and disclosed constructive baselines, this work aims to provide a reproducible reference point for execution-aware vehicle assistants rather than a claim of superiority over rules.

% ---------------------------------------------------------------------
\paragraph{Reproducibility.} The code, checkpoint, dataset, and logs are not submitted with this paper. Implementation and training details are withheld for commercial reasons.

% ---------------------------------------------------------------------
\appendix
\section{Parameter accounting}
\label{app:params}
\begin{table}[h]
\centering
\small
\begin{tabular}{lr}
\toprule
component & parameters\\
\midrule
\mbert{} encoder (22 layers, hidden 768, vocab 256k) & 306{,}939{,}648\\
existing typed decision heads (choice/scorer/act) & 14{,}175{,}744\\
GroupCalibration relation head & 1{,}611{,}868\\
other calibration / embeddings / buffers & 594{,}606\\
\midrule
full checkpoint & 323{,}520{,}866\\
trainable in joint fine-tune (tail 4 layers + norm + head) & 21{,}672{,}796\\
\bottomrule
\end{tabular}
\end{table}


\begin{thebibliography}{99}
\bibitem{modernbert}
Benjamin Warner, Antoine Chaffin, Benjamin Clavi\'e, et al.
\newblock Smarter, better, faster, longer: A modern bidirectional encoder for fast, memory efficient, and long context finetuning and inference.
\newblock arXiv:2412.13663, 2024. \url{https://arxiv.org/abs/2412.13663}

\bibitem{mmBERT}
JHU-CLSP.
\newblock mmBERT: ModernBERT goes multilingual.
\newblock Hugging Face model card, 2025. \url{https://huggingface.co/jhu-clsp/mmBERT-base}

\bibitem{focal}
Tsung-Yi Lin, Priya Goyal, Ross Girshick, Kaiming He, and Piotr Doll\'ar.
\newblock Focal loss for dense object detection.
\newblock In \emph{Proceedings of the IEEE International Conference on Computer Vision (ICCV)}, pages 2980--2988, 2017.

\bibitem{guo2017calibration}
Chuan Guo, Geoff Pleiss, Yu Sun, and Kilian Q. Weinberger.
\newblock On calibration of modern neural networks.
\newblock In \emph{Proceedings of the 34th International Conference on Machine Learning (ICML)}, pages 1321--1330, 2017.

\bibitem{glgin}
Libo Qin, Fuxuan Wei, Tianbao Xie, Xiao Xu, Wanxiang Che, and Ting Liu.
\newblock GL-GIN: Fast and accurate non-autoregressive model for joint multiple intent detection and slot filling.
\newblock In \emph{Proceedings of the 59th Annual Meeting of the ACL and the 11th IJCNLP (Volume 1: Long Papers)}, pages 178--188, 2021. \url{https://aclanthology.org/2021.acl-long.15/}

\bibitem{agif}
Libo Qin, Xiao Xu, Wanxiang Che, and Ting Liu.
\newblock AGIF: An adaptive graph-interactive framework for joint multiple intent detection and slot filling.
\newblock In \emph{Findings of the Association for Computational Linguistics: EMNLP 2020}, pages 1807--1816, 2020. \url{https://aclanthology.org/2020.findings-emnlp.163/}

\bibitem{neurosymbolic}
Tarek R. Besold, Artur d'Avila Garcez, Sebastian Bader, et al.
\newblock Neural-symbolic learning and reasoning: A survey and interpretation.
\newblock arXiv:1711.03902, 2017. \url{https://arxiv.org/abs/1711.03902}

\bibitem{gangadharaiah2019}
Rashmi Gangadharaiah and Balakrishnan Narayanaswamy.
\newblock Joint multiple intent detection and slot labeling for goal-oriented dialog.
\newblock In \emph{Proceedings of the 2019 Conference of the North American Chapter of the Association for Computational Linguistics: Human Language Technologies, Volume 1 (Long and Short Papers)}, pages 564--569, 2019. \url{https://aclanthology.org/N19-1055/}

\bibitem{snips}
Alice Coucke, Alaa Saade, Adrien Ball, Th\'eodore Bluche, Alexandre Caulier, David Leroy, Cl\'ement Doumouro, Thibault Gisselbrecht, Francesco Caltagirone, Thibaut Lavril, Ma\"el Primet, and Joseph Dureau.
\newblock Snips voice platform: An embedded spoken language understanding system for private-by-design voice interfaces.
\newblock arXiv:1805.10190, 2018. \url{https://arxiv.org/abs/1805.10190}

\bibitem{matres}
Qiang Ning, Hao Wu, and Dan Roth.
\newblock A multi-axis annotation scheme for event temporal relations.
\newblock In \emph{Proceedings of the 56th Annual Meeting of the Association for Computational Linguistics (Volume 1: Long Papers)}, pages 1318--1328, 2018. \url{https://aclanthology.org/P18-1122/}

\bibitem{react}
Shunyu Yao, Jeffrey Zhao, Dian Yu, Nan Du, Izhak Shafran, Karthik Narasimhan, and Yuan Cao.
\newblock ReAct: Synergizing reasoning and acting in language models.
\newblock In \emph{Proceedings of the Eleventh International Conference on Learning Representations (ICLR)}, 2023. \url{https://arxiv.org/abs/2210.03629}

\bibitem{hugginggpt}
Yongliang Shen, Kaitao Song, Xu Tan, Dongsheng Li, Weiming Lu, and Yueting Zhuang.
\newblock HuggingGPT: Solving AI tasks with ChatGPT and its friends in Hugging Face.
\newblock In \emph{Advances in Neural Information Processing Systems 36 (NeurIPS)}, 2023. \url{https://arxiv.org/abs/2303.17580}

\bibitem{plansolve}
Lei Wang, Wanyu Xu, Yihuai Lan, Zhiqiang Hu, Yunshi Lan, Roy Ka-Wei Lee, and Ee-Peng Lim.
\newblock Plan-and-solve prompting: Improving zero-shot chain-of-thought reasoning by large language models.
\newblock In \emph{Proceedings of the 61st Annual Meeting of the Association for Computational Linguistics (Volume 1: Long Papers)}, pages 2609--2634, 2023. \url{https://aclanthology.org/2023.acl-long.147/}
\end{thebibliography}
\end{document}